\documentclass[11pt,a4paper]{article}
\usepackage[hyperref]{emnlp2020}
\usepackage{times}
\usepackage{latexsym}

\usepackage{amsmath}
\usepackage{url}
\DeclareMathOperator{\arccosh}{arccosh}
\usepackage{graphicx}
\usepackage{multirow}
\usepackage{booktabs}
\usepackage{amssymb}
\usepackage{adjustbox}
\usepackage{makecell}
\usepackage{diagbox}
\usepackage{xspace}
\usepackage{xcolor}
\usepackage{colortbl,booktabs}

\newenvironment{noindlist}{
\begin{list}{\labelitemi}{
\leftmargin=1.0em 
\itemindent=0em 
\itemsep=2pt 
\parsep=1pt 
\parskip=1pt
}}{\end{list}}

\usepackage{microtype}

\aclfinalcopy 

\newcommand{\modelname}{HyperKA\xspace}

\title{Knowledge Association with Hyperbolic Knowledge Graph Embeddings}

\author{Zequn Sun$^1$, 
 		Muhao Chen$^{2,3}$, 
 		Wei Hu$^1$, 
 		Chengming Wang$^1$, 
 		Jian Dai$^4$, 
 		Wei Zhang$^4$\\
  $^1$State Key Laboratory for Novel Software Technology, Nanjing University, China \\
  $^2$Department of Computer and Information Science, University of Pennsylvania, USA \\
  $^3$Information Sciences Institute, University of Southern California, USA \\
  $^4$Alibaba Group, China \\
  \normalsize\texttt{zqsun.nju@gmail.com,\,muhaoche@usc.edu,\,whu@nju.edu.cn} \\
  \normalsize\texttt{cmwang.nju@gmail.com,\,\{yiding.dj,lantu.zw\}@alibaba-inc.com}}

\date{}

\begin{document}
\maketitle

\begin{abstract}
Capturing associations for knowledge graphs (KGs) through entity alignment, entity type inference and other related tasks benefits NLP applications with comprehensive knowledge representations. Recent related methods built on Euclidean embeddings are challenged by the hierarchical structures and different scales of KGs. They also depend on high embedding dimensions to realize enough expressiveness. Differently, we explore with low-dimensional hyperbolic embeddings for knowledge association. We propose a hyperbolic relational graph neural network for KG embedding and capture knowledge associations with a hyperbolic transformation. Extensive experiments on entity alignment and type inference demonstrate the effectiveness and efficiency of our method.
\end{abstract}

\section{Introduction}
Knowledge graphs (KGs) have emerged as the driving force of many NLP applications, e.g., KBQA \cite{HixonCH15}, dialogue generation \cite{OpenDialKG} and narrative prediction \cite{chen2019incorporating}. Different KGs are usually extracted from separate data sources or contributed by people with different expertise. Therefore, it is natural for these KGs to constitute complementary knowledge of the world that can be expressed in different languages, structures and levels of specificity~\cite{lehmann2015dbpedia,speer2017conceptnet}. Associating multiple KGs via entity alignment \cite{MTransE} or type inference \cite{JOIE} particularly provides downstream applications with more comprehensive knowledge representations. 

\begin{figure}[!t]
	\centering 
	\includegraphics[width=0.98\linewidth]{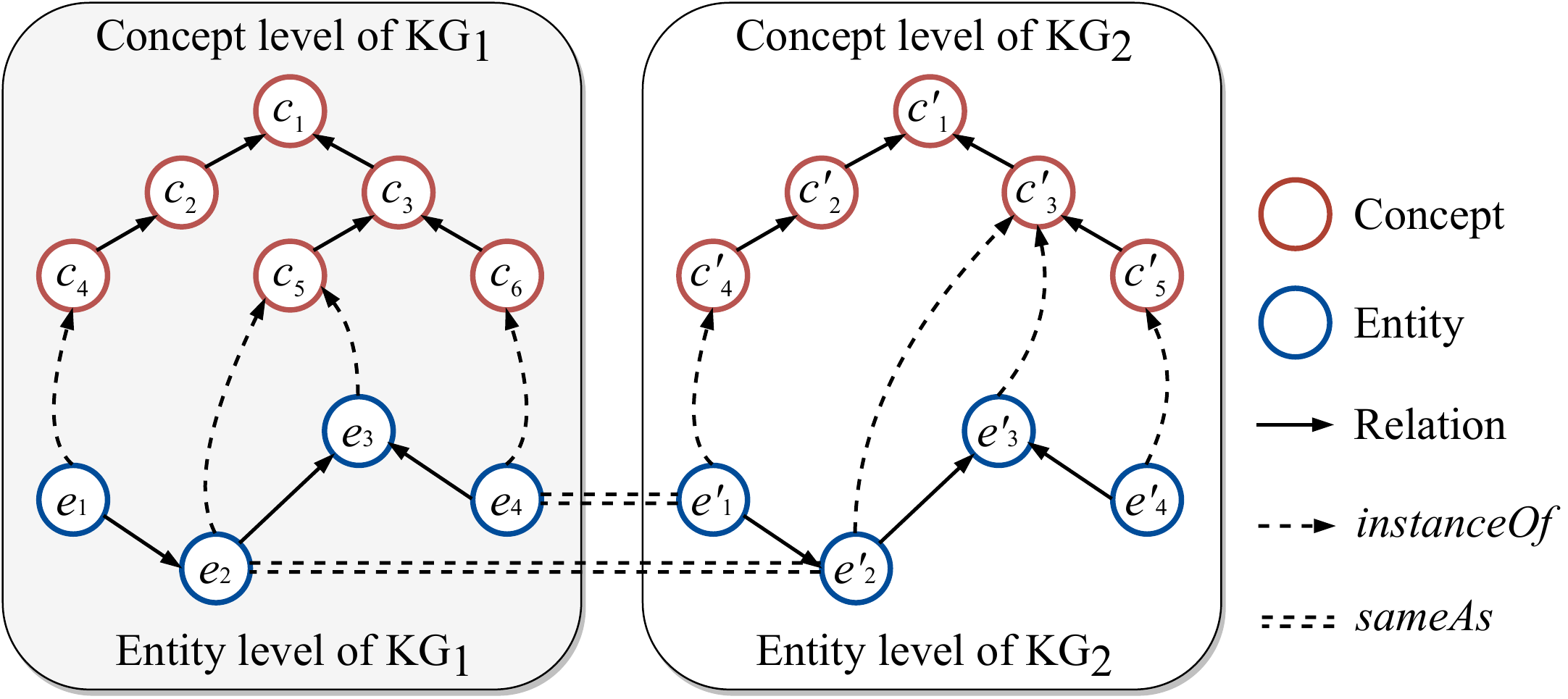}
	\caption{Illustration of two kinds of knowledge associations (i.e., \textit{sameAs} and \textit{instanceOf}) in KGs.}\label{fig:example}
	\vspace{-1.2em}
\end{figure}

Entity alignment and type inference seek to find two kinds of knowledge associations, i.e., \textit{sameAs} and \textit{instanceOf}, respectively. An example showing such associations is given in Figure \ref{fig:example}. Specifically, \emph{entity alignment} is to find equivalent entities from different entity-level KGs, such as \textit{United States} in DBpedia and \textit{United States of America} in Wikidata. \emph{Type inference}, on the other hand, associates a specific entity with a concept describing its type information, such as \textit{United States} and \textit{Country}. The main difference lies in whether such knowledge associations express the same level of specificity or not. Challenged by the diverse schemata, relational structures and granularities of knowledge representations in different KGs \cite{Schema_Heterogeneity}, traditional symbolic methods usually fall short of supporting heterogeneous knowledge association \cite{PARIS,SIGMa,Type_Inference}. Recently, increasing efforts have been put into exploring embedding-based methods \cite{MTransE,LinkNBed,Entity_Typing}. Such methods capture the associations of entities or concepts in a vector space, which can help overcome the symbolic and schematic heterogeneity \cite{JAPE}.

Embedding-based knowledge association methods still face challenges in the following aspects. (\romannumeral1) \textit{Hierarchical structures.} A KG usually consists of many local hierarchical structures \cite{EntityHierarchy15}. Besides, a KG also usually comes with an ontology to manage the relations (e.g., \textit{subClassOf}) of concepts \cite{JOIE}, which typically forms hierarchical structures as illustrated in Figure~\ref{fig:example}. It is particularly difficult to preserve such hierarchical structures in a linear embedding space \cite{nickel2014reducing}. (\romannumeral2) \textit{High parameter complexity.} To enhance the expressiveness of KG embeddings, many methods require high embedding dimensions, which inevitably cause excessive memory consumption and intractable parameter complexity. For example, for the entity alignment method GCN-Align \cite{GCN_Align}, the embedding dimension is selected to be as large as $1,000$. Reducing the dimensions can effectively decrease memory cost and training time. (\romannumeral3) \textit{Different scales.} The KGs that we manipulate may differ in scales. For example, while the English DBpedia contains $4,233,000$ entities, its ontology only contains less than a thousand concepts. Capturing the associations between entities and concepts has to deal with drastically different scales of structures and search spaces, while most existing methods do not consider such difference.

To tackle these challenges, we propose a novel hyperbolic knowledge association method, namely \modelname, inspired by the recent success of hyperbolic representation learning \cite{NickelK17,HyperText,HyperWord}. Unlike the Euclidean circle circumference that grows linearly w.r.t. the radius, the hyperbolic space grows exponentially with the radius. This property makes the hyperbolic geometry particularly suitable for embedding the hierarchical structures that drastically span their sizes along with their levels. It is also capable of achieving superior expressiveness at a low dimension. To leverage such merit, \modelname employs a hyperbolic relational graph neural network (GNN) for KG embedding and captures multi-granular knowledge associations with a hyperbolic transformation between embedding spaces. For each KG, \modelname first incorporates hyperbolic translational embeddings at the input layer of the GNN. Then, several hyperbolic graph convolution layers are stacked over the inputs to aggregate neighborhood information and obtain the final embeddings of entities or concepts. On top of the KG embeddings, a hyperbolic transformation is jointly trained to capture the associations. We conduct extensive experiments on entity alignment and type inference. \modelname outperforms SOTA methods on both tasks at a moderate dimension (e.g., $50$ or $75$). Even with a small dimension (e.g., $10$), our method still shows competitive performance.

\section{Background}
\label{sect:related_work}

\subsection{Knowledge Association}
Knowledge association aims at capturing the correspondence between structured knowledge that is described under the same or different specificity. In this paper, we consider two knowledge association tasks, i.e., \emph{entity alignment} between two entity-level KGs and \emph{type inference} from an entity-level KG to an ontological one. We define a KG as a 3-tuple $\mathcal{K}=\{\mathcal{E},\mathcal{R},\mathcal{T}\}$, where $\mathcal{E}$ denotes the set of objects such as entities or concepts. $\mathcal{R}$ denotes the set of relations and $\mathcal{T}\subseteq \mathcal{E}\times\mathcal{R}\times\mathcal{E}$ denotes the set of triples. Each triple $\tau=(h,r,t)$ records a relation $r$ between the head and tail objects $h$ and $t$. On top of this, the associations between two entity-level KGs (or between one entity-level and one ontological KGs) $\mathcal{K}_1=\{\mathcal{E}_1,\mathcal{R}_1,\mathcal{T}_1\}$ and $\mathcal{K}_2=\{\mathcal{E}_2,\mathcal{R}_2,\mathcal{T}_2\}$ are defined as $\mathcal{A}=\{(i,j)\in \mathcal{E}_1\times\mathcal{E}_2 \,|\,i\to j\}$, where $\to$ denotes a kind of associations, such as the \textit{sameAs} relationship for entity alignment or the \textit{instanceOf} relationship in the case of type inference. A small subset of associations $\mathcal{A}^+\subset\mathcal{A}$ are usually given as training data and we aim at finding the remaining.

\subsection{Related Work}
\noindent\textbf{Knowledge association tasks and methods.}
Entity alignment or type inference between KGs can be viewed as a knowledge association task. A typical method of entity alignment is MTransE \cite{MTransE}. It jointly conducts translational embedding learning \cite{TransE} and alignment learning to capture the matches of entities based on embedding distances or transformations. As for type inference, JOIE \cite{JOIE} deploys a similar framework to learn associations between entities and concepts. Later studies explore with three lines of techniques for improvement. (\romannumeral1) \emph{KG embedding}. Besides translational embeddings, some studies employ other relational learning techniques such as circular correlations \cite{JOIE,MMEA}, recurrent skipping networks \cite{RSN}, and adversarial learning \cite{SEA,OTEA,AKE}. Others employ various GNNs to seize the relatedness of entities based on neighborhood information, including GCN \cite{GCN_Align,MuGNN}, GAT \cite{NAEA,KECG,MRAEA} and relational GCNs \cite{wu2019relation,wu2019jointly,AliNet}. These techniques seek to better induce embeddings with more comprehensive relational modeling. Other studies for ontology embeddings \cite{TransC,Entity_Typing_JZLi} consider relative positions between spheres as the hierarchical relationships of corresponding concepts. However, they are still limited to linear embeddings, hence may easily fall short of preserving the deep hierarchical structures of KGs. (\romannumeral2) \emph{Auxiliary information}. Besides relational structures, some studies characterize entities based on auxiliary information, including numerical attributes \cite{JAPE,AttrE}, literals \cite{gesese2019survey,MultiKE} and descriptions \cite{HMAN,KDCoE,Entity_Typing}. They capture associations based on alternative resources, but are also challenged by the less availability of auxiliary information in many KGs \cite{speer2017conceptnet,mitchell2018never}. (\romannumeral3) \emph{Semi-supervised learning}. Another group of studies seek to infer associations with limited supervision, including self-learning \cite{BootEA,TransEdge,NAEA} and co-training \cite{KDCoE}. These methods are competent in inferring one-to-one entity alignment, without consideration of associations between entities and concepts. A recent survey by \citet{sun2020benchmark} has systematically summarized all three lines of studies.

\noindent\textbf{Hyperbolic representation learning.}
Different from Euclidean embeddings, some studies explore to characterize structures in hyperbolic embedding spaces, and use the non-linear hyperbolic distance to capture the relations between objects \cite{NickelK17,sala2018representation}. This technique has shown promising performance in embedding hierarchical data, e.g., co-purchase records \cite{vinh2018hyperbolic}, taxonomies \cite{le2019inferring,aly2019every} and organizational charts \cite{chen2019edge}. Further work extends hyperbolic embeddings to capture relational hierarchies of sentences \cite{HyperText}, neighborhood aggregation \cite{HGCNN,HGCN} and missing triples of a KG \cite{HyperKG,PoincareKG}. These studies mainly focus on the scenario of a single independent structure. Learning associations across multiple KG structures with hyperbolic embeddings is still an unsolved issue, which is exactly the focus of this paper.

\section{Hyperbolic Geometry}

The hyperbolic space is one of the three kinds of isotropic spaces. Table~\ref{tab:geometries} lists some key properties of the Euclidean (flat), spherical (positively curved) and hyperbolic (negatively curved) spaces. Compared with the Euclidean and spherical spaces, the amount of space covered by a hyperbolic geometry increases exponentially rather than polynomially w.r.t. the radius. This property allows us to capture KG structures at a very low dimension, and particularly suits those forming hierarchies. For the hyperbolic geometry, there are several important models including the hyperboloid model \cite{reynolds1993hyperbolic}, Klein disk model \cite{nielsen2014visualizing} and Poincar\'e ball model \cite{cannon1997hyperbolic}. In this paper, we choose the Poincar\'e ball model due to its feasibility for gradient optimization \cite{PoincareKG}. Specifically, the $n$-dimensional Poincar\'e ball with a negative curvature $-c$ ($c>0$) is defined by the manifold $\mathbb{D}^{n,c}=\{x\in \mathbb{R}^n\,|\,\|x\| < \frac{1}{c}\}$. For simplicity, we follow~\cite{HNN} and let $c=1$. We hereby introduce some basic operations of hyperbolic geometry, which we use extensively.

\begin{table}[!t]
\begin{adjustbox}{width=0.99\columnwidth}
\begin{tabular}{|l|c|c|c|}\hline
\diagbox{Property}{Geometry} & Euclidean & Spherical & Hyperbolic \\ \hline
Curvature & $0$ & $>0$ & $<0$ \\ \hline
Parallel lines & $1$ & $0$ & $\infty$ \\ \hline
\makecell*[l]{Shape of triangles} &
\makecell*[c]{\includegraphics[width=.3in]{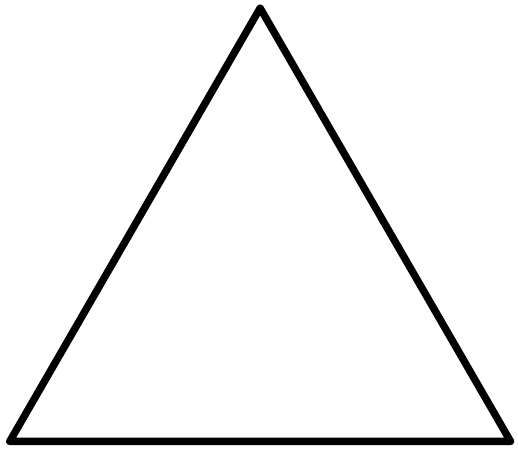}} &
\makecell*[c]{\includegraphics[width=.3in]{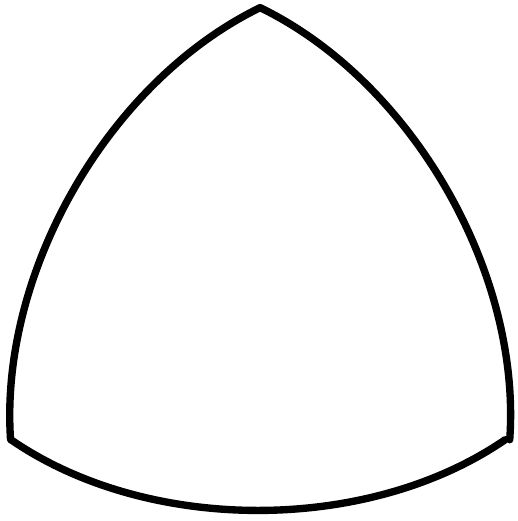}} &
\makecell*[c]{\includegraphics[width=.3in]{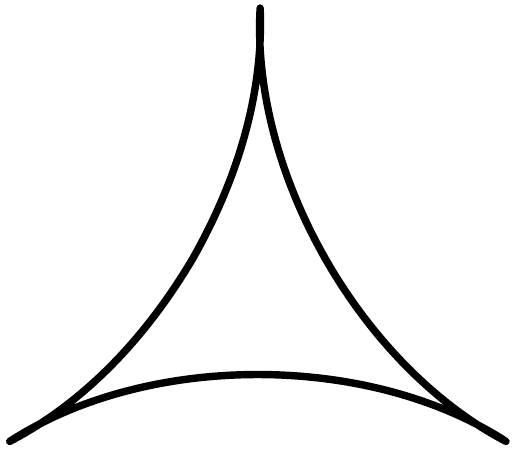}} \\ \hline
Sum of triangle angles & $\pi$ & $>\pi$ & $<\pi$ \\
\hline
\end{tabular}
\end{adjustbox}
\caption{Characteristic properties of Euclidean, spherical and
hyperbolic geometries~\cite{Hyperbolic_Geometry}. \label{tab:geometries}}
\vspace{-1.5em}
\end{table}

\noindent
\textbf{Hyperbolic distance.} The distance between vectors $\mathbf{u}$ and $\mathbf{v}$ in the Poincar\'e ball is given by:
\begin{equation*}
\label{eq:hyperbolic_distance}
  d_{\mathbb{D}}(\mathbf{u}, \mathbf{v}) = \arccosh (1+2\frac{\|\mathbf{u} - \mathbf{v}\|^2}{(1-\|\mathbf{u}\|^2)(1-\|\mathbf{v}\|^2)}).
\end{equation*}

\noindent
When points move from the origin towards the ball boundary, their distance would increase exponentially, offering a much larger volume of space for embedding learning.

\noindent
\textbf{Vector translation.} The vector translation in the Poincar\'e ball is defined by the M\"obius addition:
\begin{equation*}
\label{eq:hyperbolic_addition}
  \mathbf{u} \oplus \mathbf{v} = \frac{(1+2\langle\mathbf{u},\mathbf{v}\rangle+\|\mathbf{v}\|^2)\mathbf{u} + (1-\|\mathbf{u}\|^2)\mathbf{v}}{1+2\langle \mathbf{u},\mathbf{v}\rangle + \|\mathbf{u}\|^2\|\mathbf{v}\|^2}.
\end{equation*}

\noindent
\textbf{Transformation.} A transformation is the backbone of both GNNs \cite{HGCNN,HGCN} and transformation-based associations \cite{MTransE,JOIE}. The work in \cite{HNN} defines the matrix-vector multiplication between Poincaré balls using the exponential and logarithmic maps. The hyperbolic vectors are first projected into the tangent space at $\mathbf{0}$ using the logarithmic map ($\log_{\mathbf{0}}:\mathbb{D}^{n,1}\to T_{\mathbf{0},n}\mathbb{D}^{n,1}$), then multiplied the transformation matrix like in the Euclidean space, and finally projected back on the manifold with the exponential map ($\exp_{\mathbf{0}}:T_{\mathbf{0},n}\mathbb{D}^{n,1}\to \mathbb{D}^{n,1}$). Specifically, the two projections on vector $\mathbf{u}\in\mathbb{D}^{n,1}$ are defined as follows:
\begin{align}\label{eq:exp}
	\exp_{\mathbf{0}}(\mathbf{u}) = \tanh(\|\mathbf{u}\|)\frac{\mathbf{u}}{\|\mathbf{u}\|},\\
	\log_{\mathbf{0}}(\mathbf{u}) = \tanh^{-1}(\|\mathbf{u}\|)\frac{\mathbf{u}}{\|\mathbf{u}\|}.
\end{align}
Through such inverse projections, theoretically, we can apply any Euclidean counterpart operations on hyperbolic vectors. The transformation that maps a vector $\mathbf{u}\in\mathbb{D}^{n,c}$ into $\mathbb{D}^{m,c}$ can be done using the M\"obius version of matrix-vector multiplication:
\begin{equation}
\label{eq:multiplication}
  \mathbf{M}\otimes\mathbf{u} = \exp_{\mathbf{0}}(\mathbf{M}\log_{\mathbf{0}}(\mathbf{u})).
\end{equation}

\section{Hyperbolic Knowledge Association}
In this section, we introduce the technical details of \modelname\ --- the hyperbolic GNN-based representation learning method for knowledge association. Different from existing relational GNNs like R-GCN \cite{R-GCN}, AVR-GCN \cite{AVR-GCN} and CompGCN \cite{CompGCN} that perform a relation-specific transformation on relational neighbors before aggregation, our method models relations as translations between entity vectors at the input layer, and performs neighborhood aggregation on top of them to derive the final entity embeddings. This allows our method to benefit from both relation translation and neighborhood aggregation without increasing computation complexity.

\subsection{Hyperbolic Relation Translation}
\label{sect:translation}
Given a triple from the KG, the translational technique \cite{TransE} models a relation as a translation vector between its head and tail entities. This technique has shown promising performance on many downstream tasks such as relation prediction, triple classification and entity alignment \cite{TransE,MTransE,TransEdge}. An apparent issue of such translations to embed hierarchies in the Euclidean space is that it would require a large space to preserve the successive relation translations in a hierarchical structure. The data in a hierarchy grows exponentially w.r.t. its levels, while the amount of space grows linearly in a Euclidean space. As a result, the Euclidean embeddings usually come with a high dimension so as to achieve enough expressiveness for the aforementioned hierarchical structures. However, such modeling can be easily done in the hyperbolic space with a low dimension, where the distance between two points increases exponentially as they move towards to the boundary of the hyper-sphere. 

In our method, we seek to migrate the original translation operation to the hyperbolic space in a compact way. Accordingly, the following energy function is defined for a triple $\tau=(h,r,t)$:
\begin{equation}\label{relation_score}
  f(\tau)=d_{\mathbb{D}}(\mathbf{u}_h^{(0)} \oplus \mathbf{u}_r^{(0)},\mathbf{u}_t^{(0)}),
\end{equation}
where $\mathbf{u}_h^{(0)},\mathbf{u}_r^{(0)},\mathbf{u}_t^{(0)}\in \mathbb{D}^{n,c}$ denote the embeddings for $h$, $r$ and $t$ at the input layer, respectively. Our method is different from some existing methods \cite{PoincareKG,HyperKG} that use hyperbolic relation-specific transformations on entity representations and may easily cause high complexity overhead. The parameter complexity of our translation operation remains the same as TransE. We prefer low energy for positive triples while high energy for negatives. Hence, we minimize the following contrastive learning loss:
\begin{equation}
  \label{eq:hyperTransE_loss}
  \mathcal{L}_{\text{rel}}=\sum_{\tau\in\mathcal{T}_1\cup\mathcal{T}_2}f(\tau) + \sum_{\tau'\in\mathcal{T}^-}[\lambda_1 - f(\tau')]_+,
\end{equation}
where $\mathcal{T}^-$ denotes the set of negative triples generated by corrupting positive triples \cite{BootEA}. $\lambda_1$ is the margin where we expect $f(\tau')>\lambda_1$. 

\subsection{Hyperbolic Neighborhood Aggregation}
GNNs \cite{GCN} have recently become the paradigm for graph representation learning. Particularly, for the entity alignment task, the main merit of GNN-based methods lies in capturing the high-order proximity of entities based on their neighborhood information \cite{GCN_Align}. Inspired by the recent proposal of hyperbolic GNNs \cite{HGCN,HGCNN}, we seek to use the hyperbolic graph convolution to learn embeddings for knowledge association. The typical message passing process of GNNs consists of two phases, i.e., \underline{agg}regating neighborhood features
\begin{equation}
  \label{eq:aggregation}
  \mathbf{u}^{(l)}_{\mathcal{N}(i)} = \text{agg}(\{\mathbf{u}^{(l-1)}_{j} | j\in\mathcal{N}(i)\}),
\end{equation}
and \underline{comb}ining node and neighborhood information
\begin{equation}
  \label{eq:combination}
  \mathbf{u}^{(l)}_i = \text{comb}(\mathbf{u}^{(l-1)}_i,\mathbf{u}^{(l)}_{\mathcal{N}(i)}),
\end{equation}
where $\mathbf{u}^{(l)}_{\mathcal{N}(i)}$ denotes the representation of central object $i$ by aggregating its neighborhood information $\mathcal{N}(i)$ at the $l$-th layer. $\mathbf{u}^{(l)}_i$ denotes the representation of object $i$ by combining its representation from the last layer $\mathbf{u}^{(l-1)}_i$ and the aggregated representation of its neighborhood $\mathbf{u}^{(l)}_{\mathcal{N}(i)}$. 

Different aggregation and combination functions lead to different variants of GNNs. We choose the message passing technique that highlights the representations of central objects, to benefit from the translational embeddings at the input layer. Specifically, the message passing process of our hyperbolic GNN from the $(l-1)$-th layer to the $l$-th layer is defined as follows:
\begin{equation} 
	\label{eq:gcn}
	  \mathbf{u}^{(l)}_i = \mathbf{u}^{(l-1)}_i\oplus \sigma(\mathbf{W}^{(l)}\otimes\mathbf{u}^{(l-1)}_{\mathcal{N}'(i)}\oplus \mathbf{b}^{(l)}),
\end{equation}

\noindent
where $\mathbf{W}^{(l)}$ is the transformation matrix and $\mathbf{b}^{(l)}$ is the bias vector at the $l$-th layer. $\sigma$ is an activation function. We adopt mean-pooling to compute $\mathbf{u}^{(l-1)}_{\mathcal{N}'(i)}$ based on the representations of entity $i$ and its neighbors from the $(l-1)$-th layer. Generally, we can use the output representation of the final layer $\mathbf{u}_i=\mathbf{u}_i^{(L)}$ to represent object $i$, where $L$ is the number of GNN layers. To further benefit from relation translation, we can also combine the input and output representations $\mathbf{u}_i=\mathbf{u}_i^{(0)}\oplus\mathbf{u}_i^{(L)}$ as the final embeddings for knowledge association.

\subsection{Hyperbolic Knowledge Projection}
Once each KG is embedded in a hyperbolic space, the next step is to capture the associations between different KGs. Many previous studies jointly embed different KGs into a unified space \cite{JAPE,GCN_Align,KECG}, and infer the associations based on similarity of entity embeddings. However, pursing similar embeddings in a shared space is ill-posed for KGs with inconsistent structures, especially under the cases with different scales of knowledge representations. We hereby tackle the challenge with a knowledge projection technique in the hyperbolic space. Given a pair of seed knowledge association $(i,j)\in\mathcal{A}^+$, we use the M\"obius multiplication to project $\mathbf{u}_i$ to find the target $\mathbf{u}_j$ in the other space. The transformation error is defined as the hyperbolic distance between projected embeddings:
\begin{equation}
  \label{eq:proj_error}
\pi(i,j) = d_{\mathbb{D}}(\mathbf{M}\otimes\mathbf{u}_i,\mathbf{u}_j),
\end{equation}
where $\mathbf{M}\in \mathbb{R}^{n\times m}$ serves as the linear transformation from the hyperbolic space $\mathbb{D}^{n,c}$ of $\mathcal{K}_1$ to $\mathbb{D}^{m,c}$ of $\mathcal{K}_2$. The two hyperbolic spaces are not necessarily of the same dimension, i.e., we usually have $n\neq m$. The projection loss is defined as follows:
\begin{equation}
  \label{eq:proj_loss}\small
  \mathcal{L}_{\text{proj}}=\sum_{(i,j)\in\mathcal{A}^+}\pi(i,j)  + \sum_{(i',j')\in\mathcal{A}^{-}}[\lambda_2-\pi(i',j')]_+,
\end{equation}
$\mathcal{A}^{-}$ thereof is the set of negative samples of knowledge associations, and $\lambda_2>0$ is a margin. 

\subsection{Training}
The overall loss of the proposed method is the combination of relation translation learning and knowledge projection learning, which is given by:
\begin{equation}
  \label{eq:loss}
  \mathcal{L} =\mathcal{L}_{\text{rel}} + \mathcal{L}_{\text{proj}}.
\end{equation}

\noindent
The embedding vectors are initialized using the Xavier normal initializer. Then, we can use the exponential map to project vectors to the Poincar\'e ball. We adopt the Riemannian SGD algorithm \cite{Bonnabel13RSGD} to optimize the loss function. Let $\mathbf{\theta}$ be the trainable parameters. The Riemannian gradient $\nabla_{\text{H}}$ at $\mathbf{\theta}^t$ is computed as follows:
\begin{equation}
  \label{eq:gradient}
 \nabla_{\text{H}}= \frac{(1-\|\theta^t\|^2)^2}{4} \nabla_{\text{E}},
\end{equation}
where $\nabla_{\text{E}}$ denotes the Euclidean gradient. We use Adam \cite{Adam} as the optimizer.

\section{Experiments}
We evaluate the proposed method \modelname on two tasks of knowledge association, i.e. entity alignment (Section \ref{subsect:ea}) and entity type inference (Section \ref{subsect:ti}). The source code is publicly available\footnote{\url{https://github.com/nju-websoft/HyperKA}}.

\begin{table*}[!t]
\centering
\resizebox{0.999\textwidth}{!}{
\begin{tabular}{lllcccccccc}
	\toprule
	\multirow{2}{*}{Methods} & \multirow{2}{*}{Dimensions} & \multicolumn{3}{c}{ZH-EN} & \multicolumn{3}{c}{JA-EN} & \multicolumn{3}{c}{FR-EN} \\
	\cmidrule(lr){3-5} \cmidrule(lr){6-8} \cmidrule(lr){9-11}
	& & H@1 & H@10 & MRR & H@1 & H@10 & MRR & H@1 & H@10 & MRR \\ 
	\midrule
	MTransE \cite{MTransE} & 75 & 0.308 & 0.614 & 0.364 & 0.279 & 0.575 & 0.349 & 0.244 & 0.556 & 0.335 \\
	IPTransE \cite{IPTransE} & 75 & 0.406 & 0.735 & 0.516 & 0.367 & 0.693 & 0.474 & 0.333 & 0.685 & 0.451 \\
	AlignE~\cite{BootEA} & 75 & 0.472 & 0.792 & 0.581 & 0.448 & 0.789 & 0.563 & 0.481 & 0.824 & 0.599 \\
	SEA~\cite{SEA} & 75 & 0.424 & 0.796 & 0.548 & 0.385 & 0.783 & 0.518 & 0.400 & 0.797 & 0.533 \\
	RSN4EA~\cite{RSN} & 300 & 0.508 & 0.745 & 0.591 & 0.507 & 0.737 & 0.590 & 0.516 & 0.768 & 0.605 \\
	\midrule
	GCN-Align~\cite{GCN_Align} & 1000, 1000, 1000 & 0.413& 0.744 & 0.549 & 0.399 & 0.745 & 0.546 & 0.373 & 0.745 & 0.532 \\
	MuGNN~\cite{MuGNN} & 128, 128, 128 & 0.494 & \underline{0.844} & 0.611 & 0.501 & \underline{0.857} & 0.621 & 0.495 & \underline{0.870} & 0.621 \\
	KECG~\cite{KECG} & 128, 128, 128, 128 & 0.478 & 0.835 & 0.598 &  0.490 & 0.844 & 0.610 & 0.486 & 0.851 & 0.610 \\
	AliNet~\cite{AliNet} & 500, 400, 300 & \underline{0.539} & 0.826 & \underline{0.628} & \underline{0.549} & 0.831 & \underline{0.645} & \underline{0.552} & 0.852 & \underline{0.657}\\
	\midrule
	\modelname (w/o relation) & 75, 75, 75 & 0.518 & 0.814 & 0.623 & 0.535 & 0.834 & 0.640 & 0.529 & 0.859 & 0.645\\
	\modelname & 75, 75, 75 & \textbf{0.572} & \textbf{0.865} & \textbf{0.678} & \textbf{0.564} & \textbf{0.865} & \textbf{0.673} & \textbf{0.597} & \textbf{0.891} & \textbf{0.704}\\
	\bottomrule
\end{tabular}}
\caption{Entity alignment results on DBP15K. For the dimension of GNN-based methods, we report the output dimensions of their input layer and GNN layers. The best scores are in bold and the second-best ones are underlined.}
\label{tab:ent_alignment}
\end{table*}

\subsection{Entity Alignment}
\label{subsect:ea}
Entity alignment aims at matching the counterpart entities that describe the same real-world identity across two entity-level KGs.
The inference of entity alignment is based on the embedding distances.

\subsubsection{Experimental Setup}
\noindent \textbf{Datasets.}
We use the widely-adopted entity alignment dataset DBP15K~\cite{JAPE} for evaluation. It is extracted from DBpedia \cite{lehmann2015dbpedia} and consists of three settings, namely ZH-EN (Chinese-English), JA-EN (Japanese-English) and FR-EN (French-English). Each setting contains 15 thousand pairs of entity alignment. The dataset splits are consistent with those in previous studies \cite{JAPE,BootEA}, which result in 30\% of entity alignment being used in training. The statistics of DBP15K are reported in Appendix A.


\noindent \textbf{Baselines.}
We compare \modelname with nine recent structure-based entity alignment methods, including five relation-based methods, i.e., MTransE \cite{MTransE}, IPTransE \cite{IPTransE}, AlignE \cite{BootEA}, SEA \cite{SEA} and RSN4EA \cite{RSN}, as well as four neighborhood-based methods, i.e., GCN-Align \cite{GCN_Align}, MuGNN \cite{MuGNN}, KECG \cite{KECG} and AliNet \cite{AliNet}. We omit here several methods that require auxiliary entity information that are not used by others (see Section \ref{sect:related_work}). We also do not involve two related methods MMEA \cite{MMEA} and MRAEA \cite{MRAEA} because their bidirectional alignment setting is different from ours and other baselines. For ablation study, we evaluate a variant of our method without relation translation, i.e., \modelname (w/o relation). The main results are reported in Section~\ref{sect:main_results}. Besides, we further consider semi-supervised entity alignment methods BootEA \cite{BootEA}, NAEA \cite{NAEA} and TransEdge \cite{TransEdge} as they achieve high performance by bootstrapping from unlabeled entity pairs. We describe the implementation of the semi-supervised \modelname variant and experimental results shortly in Section~\ref{sect:semi}.

\noindent \textbf{Model configuration.} 
In the main experiment, we use two GNN layers, and set the dimension of all layers in \modelname to 75. The dimensions for the two KGs are the same, i.e., $n=m=75$. This is the smallest dimension adopted by any baseline methods. Note that, we also evaluate our method with a range of dimensions from 10 to 150, to assess its robustness. We report in Appendix B the implementation details of \modelname and the selected values for hyper-parameters, including the learning rate, the batch size, margin values $\lambda_1$ and $\lambda_2$, etc. Following convention, we report three metrics on entity alignment, i.e., H@1 (precision), H@10 (the proportion of correct alignment ranked within the top 10) and MRR (mean reciprocal rank). Higher scores of those metrics indicate better performance.


\subsubsection{Main Results}
\label{sect:main_results}
We report the entity alignment results on DBP15K in Table~\ref{tab:ent_alignment}. Note that the embedding dimension for \modelname is set to 75 (the smallest setting among baseline methods). We can observe that \modelname consistently outperforms all baseline methods on all three datasets, especially GNN-based methods. For example, on DBP15K FR-EN, the H@1 score of \modelname reaches 0.597, surpassing MuGNN by 0.102 and AliNet by 0.045, even though \modelname uses a smaller dimension than these methods. Compared against the baselines with dimension of 75, \modelname also achieves much better performance. For instance, on the ZH-EN dataset, it surpasses AlignE by 0.1 in H@1. Overall, \modelname significantly outperforms the SOTA Euclidean methods, while using the same or much smaller dimension settings. This shows that the hyperbolic embeddings have superior expressiveness than the linear embeddings. As for the comparison between two variants of \modelname, we can see that the one with relation embedding performs notably better. This demonstrates the effectiveness of incorporating relation translation into GNNs.

\begin{table}[!t]
	\centering
	\resizebox{0.85\linewidth}{!}{
	\begin{tabular}{lcc>{\columncolor[gray]{0.9}}ccc>{\columncolor[gray]{0.9}}c}
	\toprule
	\multirow{2}{*}{Datasets} & \multicolumn{6}{c}{Dimensions} \\
	\cmidrule(lr){2-7}
	& 10 & 25 & 35 & 50 & 75 & 150 \\ \midrule
	ZH-EN & 0.370 & 0.487 & 0.532 & 0.554 & 0.572 & 0.587 \\
	JA-EN & 0.391 & 0.510 & 0.551 & 0.563 & 0.564 & 0.583 \\
	FR-EN & 0.368 & 0.528 & 0.574 & 0.585 & 0.597 & 0.611 \\
	\bottomrule
	\end{tabular}}
	\caption{H@1 performance of \modelname on DBP15K using different dimensions.}
	\label{tab:results_dim}
\end{table}

\subsubsection{Analysis on Dimensions}
We further analyze the effect of different dimensions on performance and training efficiency. We report the H@1 results of different dimensions in Table \ref{tab:results_dim}. We observe that the H@1 scores of \modelname drop along with the decrease of embedding dimensions. This observation is generally in line with our expectations because a small dimension limits the expressiveness of KG embeddings. However, \modelname still exhibits satisfying performance at very small dimensions in comparison to other methods, such as under the dimensions of 10 and 25. Specifically, \modelname with 25 dimension even outperforms a number of methods in Table~\ref{tab:ent_alignment} with much higher dimensions, e.g., AlignE, GCN-Align and KECG. Note that, \modelname with 35 dimension achieves very similar results to AliNet with layer dimensions of (500, 400, 300) and also outperforms other baseline methods. \modelname with dimension of 150 establishes a new SOTA performance for structure-based entity alignment. Overall, the low-dimension hyperbolic representations of \modelname demonstrate more precise and robust inference of counterpart entities across KGs.

\begin{figure}[t]
	\centering
	\includegraphics[width=0.88\linewidth]{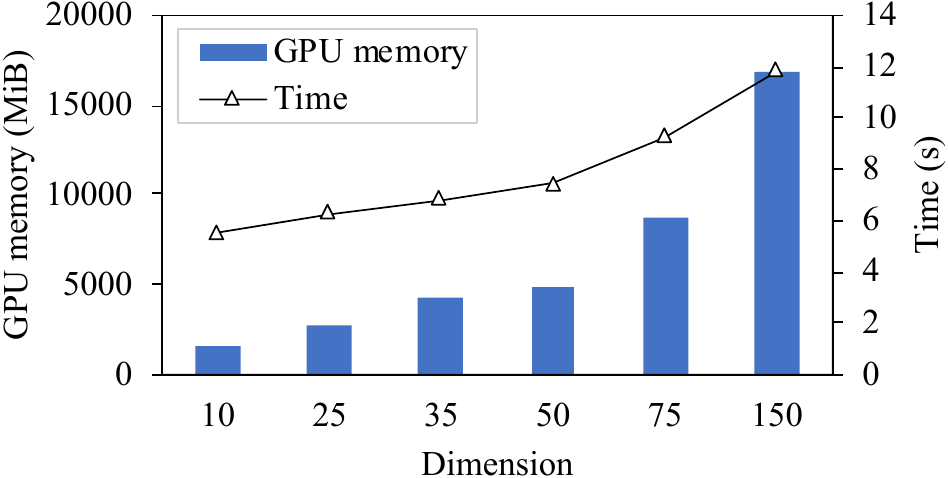}
	\caption{GPU memory cost and running time of each epoch w.r.t. different dimensions on DBP15K ZH-EN.}
	\label{fig:dim}
\end{figure}

\begin{figure}[!t]
	\centering
	\includegraphics[width=0.9\linewidth]{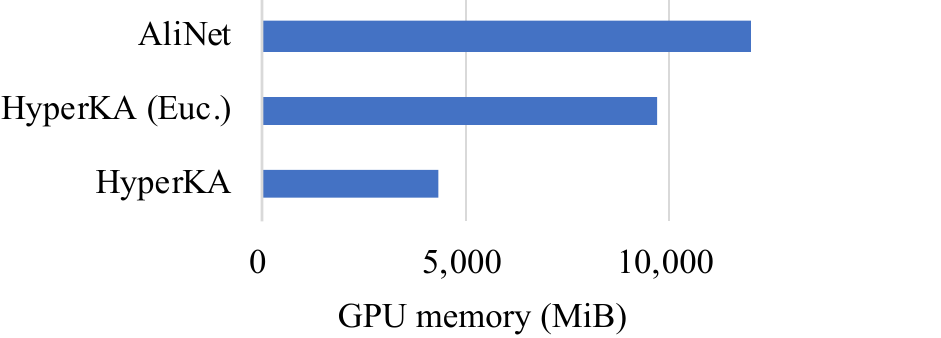}
	\caption{GPU memory cost of training \modelname and its Euclidean counterpart \modelname (Euc.) as well as AliNet~\cite{AliNet} on DBP15K ZH-EN when they achieve similar performance. The dimension settings that they need are respectively (35, 35, 35), (200, 200, 200) and (500, 400, 300), and their H@1 scores are 0.532, 0.549 and 0.539, respectively.}
	\label{fig:memory}
\end{figure}

We report in Figure \ref{fig:dim} the GPU memory costs for training \modelname in 64-bit precision settings w.r.t. various dimensions on ZH-EN, together with the average training time per epoch\footnote{The experiments are conducted on a workstation with an Intel Xeon Gold 5117 CPU and a NVIDIA Tesla V100 GPU.}. A larger dimension leads to more GPU memory costs and training time, although it also leads to better performance as shown in Table~\ref{tab:results_dim}. \modelname can achieve satisfying performance with limited GPU memory costs.

\subsubsection{Analysis on Expressiveness}
To further understand the expressiveness of our hyperbolic KG embeddings, we compare a small dimension along with their GPU memory costs of \modelname and its Euclidean counterpart \modelname (Euc.) with AliNet, when those three achieve similar performance. \modelname (Euc.) is implemented by replacing hyperbolic operations with their corresponding Euclidean operations. For example, the M\"obius addition $\oplus$ is replaced with vector addition $+$. We select the dimension of \modelname (Euc.) in $\{75, 100, 150, 200, 300, 500\}$ and its best-performing model under the dimension of 200 can achieve similar performance to AliNet. By contrast, \modelname only needs a dimension of 35 as shown in Table~\ref{tab:results_dim}. Their GPU memory costs on ZH-EN are shown in Figure~\ref{fig:memory}. We observe similar results on JA-EN and FR-EN. Specifically, \modelname only costs about 45.09\% memory of \modelname (Euc.) and 29.97\% of AliNet to achieve similar performance. This shows that hyperbolic embeddings can achieve satisfying expressiveness with a small dimension and efficient memory costs. 

\begin{table}[t]
	\centering
	\resizebox{0.81\linewidth}{!}{
	\begin{tabular}{lcccc}
	\toprule
	Datasets & Dimensions & H@1 & H@10 & MRR \\ \midrule
	ZH-EN & 200, 200, 200 & 0.549 & 0.827 & 0.650 \\
	JA-EN & 200, 200, 200 & 0.527 & 0.813 & 0.631 \\
	FR-EN & 200, 200, 200 & 0.567 & 0.864 & 0.675 \\ 
	\midrule
	ZH-EN & 300, 300, 300 & 0.581 & 0.857 & 0.683 \\
	JA-EN & 300, 300, 300 & 0.563 & 0.844 & 0.666 \\
	FR-EN & 300, 300, 300 & 0.605 & 0.896 & 0.711 \\
	\bottomrule
	\end{tabular}}
	\caption{Entity alignment results of \modelname (Euc.).}
	\label{tab:results_euc}
\end{table}

\begin{table}[!t]
\centering
\Large
\resizebox{\columnwidth}{!}{
	\begin{tabular}{lcccccc}
		\toprule
		\multirow{2}{*}{Methods} & \multicolumn{2}{c}{ZH-EN} & \multicolumn{2}{c}{JA-EN} & \multicolumn{2}{c}{FR-EN}\\ 
		\cmidrule(lr){2-3} \cmidrule(lr){4-5} \cmidrule(lr){6-7} & H@1 & MRR & H@1 & MRR & H@1 & MRR\\
		\midrule 
		BootEA~\cite{BootEA} & 0.629 & 0.703 & 0.622 & 0.701 & 0.653 & 0.731 \\ 
		NAEA~\cite{NAEA} & 0.650 & 0.720 & 0.641 & 0.718 & 0.673 & 0.752 \\
		TransEdge~\cite{TransEdge} & 0.735 & 0.801 & 0.719 & \textbf{0.795} & 0.710 & 0.796 \\
		\midrule
		\modelname (semi) & \textbf{0.743} & \textbf{0.805} & \textbf{0.727} & 0.793 & \textbf{0.741} & \textbf{0.813}\\
		\bottomrule
	\end{tabular}}
\caption{H@1 and MRR results of semi-supervised entity alignment on DBP15K. Their dimension is 75.}
\label{tab:semi}
\end{table}

\begin{table*}[t]
\centering
\resizebox{0.76\linewidth}{!}{
\begin{tabular}{lcccccccc}
	\toprule
	\multirow{2}{*}{Methods}& \multicolumn{2}{c}{Dimensions} & \multicolumn{3}{c}{{YAGO26K-906}} & \multicolumn{3}{c}{{DB111K-174}}\\
	\cmidrule(lr){2-3}
	\cmidrule(lr){4-6} \cmidrule(lr){7-9}
	& Entity & Concept & H@1 & H@3 & MRR & H@1 & H@3 & MRR\\ 
	\midrule
	TransE~\cite{TransE} & 300 & 50 & 0.732 & 0.353 & 0.144 & 0.437 & 0.608 & 0.503\\
	DistMult~\cite{DistMult} & 300 & 50 & 0.361 & 0.553 & 0.411 & 0.498 & 0.680 & 0.551\\
	HolE~\cite{HolE} & 300 & 50 & 0.348 & 0.548 & 0.395 & 0.448 & 0.654 & 0.504 \\
	\hline
	MTransE~\cite{MTransE}  & 300 & 50 & 0.609 & 0.776 & 0.689 & 0.599 & 0.813 & 0.672 \\
    JOIE~\cite{JOIE}  & 300 & 50 & 0.856 & \textbf{0.959} & 0.897 & 0.756 & \textbf{0.959} & \underline{0.857} \\
    \hline
    \modelname & 75 & 15 & \underline{0.863} & 0.946 & \underline{0.908} & \underline{0.778} & 0.918 & 0.854\\
    \modelname & 150 & 30 & \textbf{0.871} & \underline{0.948} & \textbf{0.913} & \textbf{0.789} & \underline{0.927} & \textbf{0.863} \\
	\bottomrule
\end{tabular}}
\caption{Type inference results on YAGO26K-906 and DB111K-174.}\label{tab:type inference}
\end{table*}

We report in Table~\ref{tab:results_euc} the entity alignment results of \modelname (Euc.) on DBP15K. We can find that \modelname (Euc.) with a high dimension (e.g., 300) can also achieve similar performance with \modelname at a low dimension of 75. This is because the Euclidean embeddings also have enough expressiveness to represent hierarchical structures if given a large dimension. However, hyperbolic embeddings only need a small dimension, bringing along the substantial advantage in saving memory.

\subsubsection{Semi-supervised Entity Alignment}
\label{sect:semi}
Semi-supervised entity alignment methods use self-training or co-training techniques to augment training data by iteratively finding new alignment labels~\cite{BootEA,NAEA,TransEdge}. Following BootEA~\cite{BootEA}, we use the self-training strategy to iteratively propose more aligned entity pairs to augment training data, denoted as $\mathcal{A}'=\{(i,j)\in\mathcal{E}_1 \times\mathcal{E}_2 \,|\,\pi(i, j) < \epsilon \}$, where $\epsilon$ is a distance threshold. As these pairs inevitably contains errors~\cite{BootEA}, we apply a small weight $\mu$ when using such proposed data for training, resulting in the following loss:
\begin{equation}
  \label{eq:semi_loss}
  \mathcal{L}_{\text{semi}}=\mu \sum_{(i,j)\in\mathcal{A}'}\pi(i,j).
\end{equation}

Accordingly, the semi-supervised \modelname variant minimizes the joint loss $\mathcal{L}+\mathcal{L}_{\text{semi}}$. The selected settings are $\epsilon=0.25$, $\mu=0.05$, and the training takes 800 epochs. Table~\ref{tab:semi} lists the H@1 and MRR results, where \modelname shows drastic improvement over BootEA and NAEA. It also achieves noticeably better H@1 than the latest semi-supervised method TransEdge, especially on the FR-EN setting. The good performance of TransEdge comes with prohibitive memory overhead. Its parameter complexity is $O(2N_en+N_rn)$ \cite{TransEdge}, where $N_e$ and $N_r$ denote the numbers of entities and relations in KGs, respectively. $n$ is the dimension. By contrast, the complexity of our method is $O(N_en+N_rn+Ln^2)$ and we have $N_e \gg Ln$ in practice, where $L$ is the number of GNN layers. In this case, \modelname outclasses TransEdge in both effectiveness and efficiency. Compared with our results in Table~\ref{tab:ent_alignment}, we find that the self-training, being an optional and compatible technique, brings an improvement of more than 0.14 on H@1.

\begin{figure}[!t]
	\centering
	\includegraphics[width=0.68\linewidth]{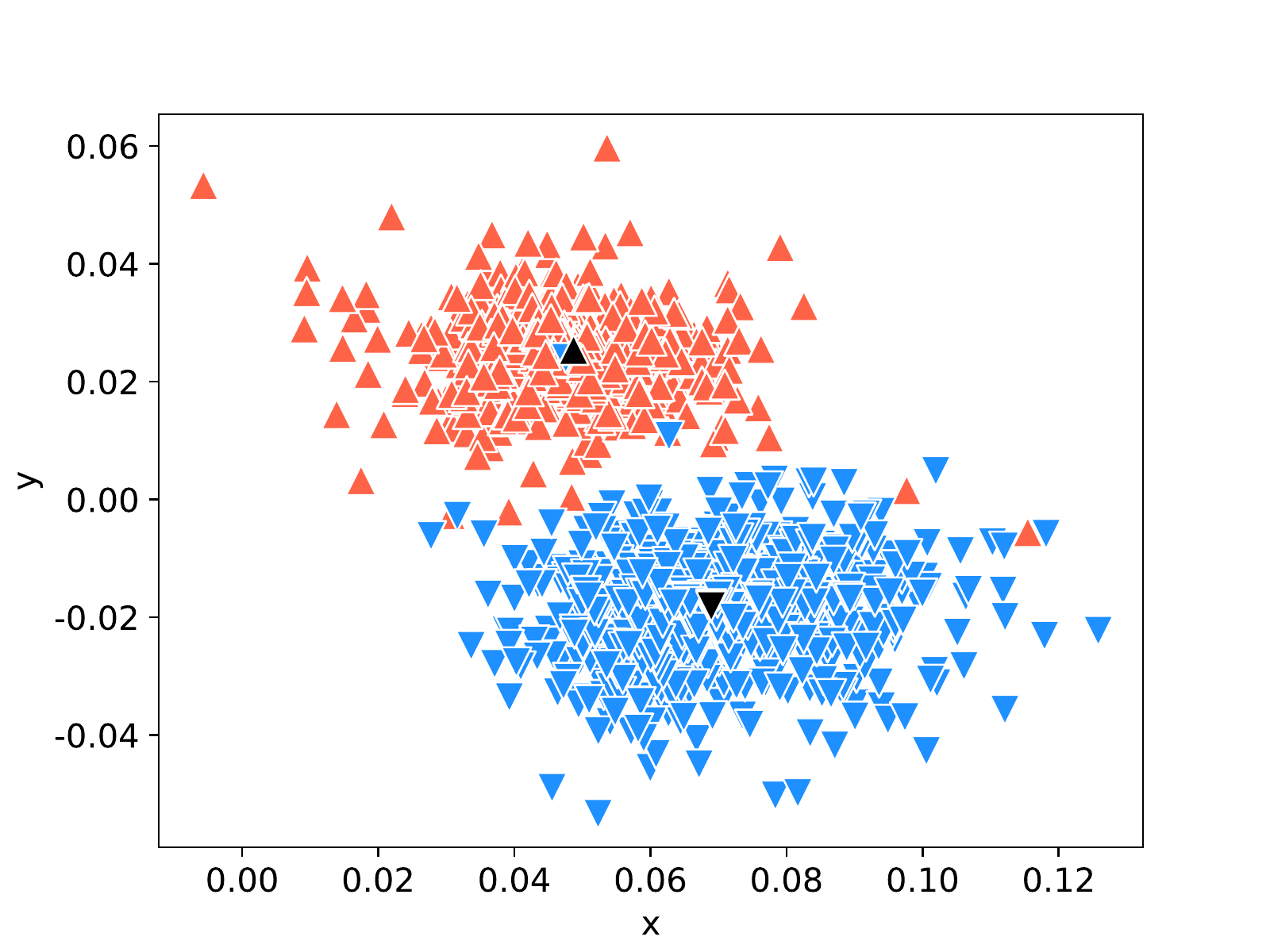}
	\caption{Visualization of the embeddings generated by \modelname for two related concepts ``\textit{Film}'' and ``\textit{Album}'' along with their entities in DB111K-174. The black up triangle denotes ``\textit{Film}'' and the surrounding red ones are its entities. The black down triangle denotes ``\textit{Album}'' and the blue ones are its entities.}\label{fig:viz}
\end{figure}

\subsection{Type Inference}
\label{subsect:ti}
The main difference between type inference and entity alignment lies in that the knowledge to associate in the former scenario differs much in scales and specificity. This causes many related methods based on shared embedding spaces to fall short. 

\subsubsection{Experimental Setup}
\noindent \textbf{Datasets}.
The experiments for this task are conducted on datasets YAGO26K-906 and DB111K-174~\cite{JOIE}, which are extracted from YAGO and DBpedia, respectively. Each dataset has an entity-level KG and an ontological KG for concepts (types). Their statistics are reported in Appendix A. To compare with the previous work \cite{JOIE}, we use the original data splits, and report H@1, H@3 and MRR results. The hyper-parameter settings are listed in Appendix B.


\noindent \textbf{Baselines}.
So far, only a few methods have been applied to the type inference task in KGs. We compare with the SOTA method JOIE \cite{JOIE}, and four other baseline methods TransE \cite{TransE}, DistMult \cite{DistMult}, HolE \cite{HolE} and MTransE \cite{MTransE} that are reported in the same paper. For JOIE, we choose its best-performing variant based on the translational encoder with cross-view transformation. A related method \cite{Entity_Typing} is not taken into comparison as it requires entity attributes that are unavailable in our problem setting.


\subsubsection{Main Results}
In this task, the embedding dimensions for entities and concepts are different, i.e., $n>m$, as an entity-level KG usually contains much more entities than concepts in a related ontological (or concept-level) KG. For \modelname, we evaluate two dimension settings: $n=75, m=15$ and $n=150, m=30$. Both are much smaller than the dimensions of baseline methods. The results are reported in Table \ref{tab:type inference}. We can observe that \modelname (75, 15) outperforms JOIE in terms of H@1 on both datasets, especially on DBP111K-174, although \modelname uses a much smaller dimension. For example, the H@1 score of \modelname (75, 15) on DB111K-174 reaches 0.778, with a gain of 0.022 over JOIE in its best setting. \modelname (150, 30) achieves the best performance over H@1 and MRR. We also try the dimension setting of (300, 50), but no longer observe further improvement. We believe this is because the dimension setting (150, 30) is enough for type inference as the concept-level KG is small. Meanwhile, once we apply the same small-dimension setting (75, 15) as \modelname to baseline methods, the performance of those methods become much worse. For example, MTransE achieves no more than 0.357 in H@1 using this small dimension.

\subsubsection{Case Study}
\label{appendix:viz}
For case study, we visualize the embeddings of two related concepts ``\textit{Film}'' and ``\textit{Album}'' in DBP111K-174 along with their associated entities in the PCA-projected space in Figure \ref{fig:viz}. Despite these two groups of entities are closely relevant, the embeddings learned by \modelname are able to clearly distinguish between these two. We can see that the entities of the same type are embedded closely after transformation, while the two clusters are generally well differentiated by a clear margin (with only a few exceptions). This displays how the hyperbolic transformation is able to capture the multi-granular associations, while preserves the gap between the entities associated with different concepts.

\section{Conclusion and Future Work}
We propose a method to capture knowledge associations with a new hyperbolic GNN-based representation learning model. The proposed \modelname\ method extends translational and GNN-based techniques to hyperbolic spaces, and captures associations by a hyperbolic transformation. Our method outperforms SOTA baselines using lower embedding dimensions on both entity alignment and type inference. For future work, we plan to incorporate hyperbolic RNNs \cite{HNN} to encode auxiliary information for zero-shot entity and concept representations. Another meaningful direction is to use \modelname to infer the associations between snapshots in temporally dynamic KGs \cite{Xu2020Inductive}. We also seek to investigate the use of \modelname for cross-domain representations of biological and medical knowledge \cite{BIO-JOIE}.

\smallskip\noindent\textbf{Acknowledgments.} We thank the anonymous re- viewers for their insightful comments. This work is supported by the National Natural Science Foundation of China (Nos. 61872172 and 61772264), and the Collaborative Innovation Center of Novel Software Technology and Industrialization.
\bibliography{reference.bib}
\bibliographystyle{acl_natbib}

\clearpage
\appendix

\section{Dataset Statistics}
\label{appendix:data}
Table \ref{tab:dataset} lists the statistics of the entity alignment dataset DBP15K\footnote{\url{https://github.com/nju-websoft/JAPE}} \cite{JAPE}, as well as two type inference datasets YAGO26K-906 and DB111K-174\footnote{\url{https://github.com/JunhengH/joie-kdd19}}~\cite{JOIE}. For a fair comparison, we reuse the original splits of associations in these datasets for training and evaluation, i.e., 30\% alignment in DBP15K as well as around 60\% associations in YAGO26K-906 and DB111K-174 as training data. We can see that the two KGs of type inference datasets differs much more in terms of the scales of objects and triples than those in entity alignment datasets, which also bring along more challenges to knowledge association.
\begin{table}[!h]	
	\centering
	\resizebox{\linewidth}{!}{
	\begin{tabular}{lcl|rrrr}
		\toprule
		\multicolumn{3}{c|}{Datasets} & \#Objects & \#Rel. & \#Triples & \#Assoc. \\ \hline
		\parbox[t]{3mm}{\multirow{6}{*}{\rotatebox[origin=c]{90}{DBP15K}}} &
		\multirow{2}{*}{ZH-EN} 
		& ZH & 66,469 & 2,830 & 153,929 &\multirow{2}{*}{15,000} \\
		& & EN & 98,125 & 2,317 & 237,674 \\ \cline{2-7}
		& \multirow{2}{*}{JA-EN } 
		& JA & 65,744 & 2,043 & 164,373 &\multirow{2}{*}{15,000} \\
		& & EN & 95,680 & 2,096 & 233,319 \\ \cline{2-7}
		& \multirow{2}{*}{FR-EN} 
		& FR & 66,858 & 1,379 & 192,191 &\multirow{2}{*}{15,000} \\
		& & EN & 105,889 & 2,209 & 278,590\\	
		\hline\hline
        \multicolumn{2}{l}{\multirow{2}{*}{YAGO26K-906}}
		& Ent. & 26,078 & 34 & 390,738 & \multirow{2}{*}{9,962} \\
		& & Ont. & 906 & 30 &  8,962 \\ 
		\hline\hline
		\multicolumn{2}{l}{\multirow{2}{*}{DB111K-174}}
		& Ent. & 111,762 &  305 &  863,643 & \multirow{2}{*}{99,748} \\
		& & Ont. & 174 & 20 & 763 \\ \bottomrule
	\end{tabular}}
	\caption{\label{tab:dataset}Statistics of the datasets used in this paper.}
\end{table}

\section{Hyper-parameter Settings}
\label{appendix:setting}
In this section, we report the implementation details and hyper-parameter settings of \modelname on the two knowledge association tasks. We select each hyper-parameter setting within a wide range of values as follows:

\begin{noindlist}
  \item Learning rate: $\{0.0001, 0.0002, 0.0005, 0.001\}$
  \item Batch size: $\{2000, 5000, 10000, 20000, 50000\}$
  \item \# GNN layers: $\{1, 2, 3, 4, 5\}$
  \item \# Negative samples: $\{1, 10, 20, 30, 40, 50, 100\}$
  \item $\lambda_1$: $\{0.01, 0.02, 0.05, 0.1, 0.15, 0.2, 0.3, 0.4\}$
  \item $\lambda_2$: $\{0.01, 0.02, 0.05, 0.1, 0.15, 0.2, 0.3, 0.4\}$
\end{noindlist}

\noindent
Table~\ref{tab:settings} lists the selected hyper-parameter settings for the best-performing models (measured by H@1 scores) of \modelname with 75 dimension for entity alignment on DBP15K, as well as (75, 15) dimensions for type inference on YAGO26K-906 and DB111K-174. We use truncated negative sampling and cross-domain similarity local scaling for the entity alignment task. The training takes 800 epochs on DBP15K, 60 epochs on YAGO26K-906 and 100 epochs on DB111K-174. The activation function used in our method is $\tanh$.

\begin{table}[ht]
\centering
\resizebox{0.999\linewidth}{!}{
\begin{tabular}{lccc}
	\toprule
	Parameters & DBP15K  & YAGO26K-906 & DB111K-174\\
	\midrule
	Learning rate & 0.0002 & 0.0005 & 0.0005 \\
	Batch size & 20,000 & 2,000 & 20,000 \\
	\# GNN layers & 2 & 3 & 3 \\
	\# Neg. samples & 40 & 40 & 30 \\
	$\lambda_1$ & 0.1 & 0.2 & 0.2 \\
	$\lambda_2$ & 0.4 & 0.1 & 0.1 \\
	\bottomrule
\end{tabular}}
\caption{Selected values for hyper-parameters.}
\label{tab:settings}
\end{table}

\end{document}